\documentclass{article}
\usepackage{amssymb}
\usepackage{wrapfig}

\usepackage[preprint]{corl_2025} 

\title{Reachability Barrier Networks: Learning Hamilton-Jacobi Solutions for Smooth and Flexible Control Barrier Functions}
\usepackage{times}
\usepackage{multirow}

\newcommand{\shnote}[1]%
    {\textcolor{magenta}{[Sylvia: #1]}}

\newcommand{\tvar}{t}
\newcommand{\thor}{t_f} 
\newcommand{\state}{x} 

\newcommand{\ctrl}{u} 
\newcommand{\csig}{\mathsf{u}}

\newcommand{\dyn}{f} 

\newcommand{\cset}{\mathcal{U}}

\newcommand{\valfunc}{V} 
\newcommand{\failureset}{\mathcal{F}}
\newcommand{\safemargin}{l}
\newcommand{\traj}{\mathsf{x}}
\newcommand{\tint}{\mathbb{T}}
\newcommand{\R}{\mathbb{R}}

\newcommand{\deepcbf}{RBN} 
\newcommand{\cbvf}{\valfunc_\gamma}
\usepackage[utf8]{inputenc} 
\usepackage[T1]{fontenc}    
\usepackage{hyperref}       
\usepackage{url}            
\usepackage{booktabs}       
\usepackage{amsfonts}       
\usepackage{nicefrac}       
\usepackage{microtype}      
\usepackage{wrapfig}
\usepackage{graphicx} 
\usepackage{graphicx} 
\usepackage{algorithm}
\usepackage{algpseudocode}
\usepackage{amsmath,amssymb}
\usepackage{bbm}
\usepackage{soul}
\usepackage{microtype,wrapfig}
\usepackage{multicol}

\usepackage[export]{adjustbox}
\usepackage{makecell}
\usepackage{caption}
\usepackage{wrapfig}

\author{
   Matthew Kim$^1$~~~~William Sharpless$^{1}$~~~~Hyun Joe Jeong$^{1}$~~~~Sander Tonkens$^{1}$\\ \textbf{Somil Bansal$^{2}$}~~~~\textbf{Sylvia Herbert$^{1}$}\\
   $^1$University of California, San Diego~~~~$^2$ Stanford University\\
   \texttt{mak009@ucsd.edu}\\ 
   \vspace{-0in}
 }

\begin{document}

\maketitle

\begin{abstract}  
Recent developments in autonomous driving and robotics underscore the necessity of safety-critical controllers. 
Control barrier functions (CBFs) are a popular method for appending safety guarantees to a general control framework, but they are notoriously difficult to generate beyond low dimensions. Existing methods often yield non-differentiable or inaccurate approximations that lack integrity, and thus fail to ensure safety.
In this work, we use physics-informed neural networks (PINNs) to generate smooth approximations of CBFs by computing Hamilton-Jacobi (HJ) optimal control solutions. These reachability barrier networks (\textbf{RBNs}) avoid traditional dimensionality constraints and support the tuning of their conservativeness post-training through a parameterized discount term. To ensure robustness of the discounted solutions, we leverage conformal prediction methods to derive probabilistic safety guarantees for RBNs. 
We demonstrate that RBNs are highly accurate in low dimensions, and safer than the standard neural CBF approach in high dimensions. Namely, we showcase the RBNs in a 9D multi-vehicle collision avoidance problem where it empirically proves to be $5.5$x safer and $1.9$x less conservative than the neural CBFs, offering a promising method to synthesize CBFs for general nonlinear autonomous systems.
\end{abstract}


\section{Introduction}
Autonomous systems have revolutionized the way humans interact with technology. However, as these systems often operate in unpredictable environments, it is necessary to ensure safety and reliability. A key challenge lies in designing control mechanisms that allow autonomous systems to achieve their goals efficiently while strictly adhering to safety requirements. For example, when an autonomous vehicle detects a potential collision with another vehicle, it should be able to safely avoid the vehicle, ideally without much deviation from its original trajectory. In a different setting, a user of a robotic arm may want the arm to deliver a cup to a specific location without damaging the cup. These examples showcase why autonomous systems must be scalable and verifiably safe in order to be deployed in the real world. Toward this end, safety filters \cite{hsu2023safety}, a control-theoretic technique for providing safety guarantees to the system, are useful because they minimally modify the nominal actions of the robot while keeping it safe.

Control barrier functions (CBFs) \cite{ames2019control} are a popular way to preserve safety because they not only guarantee safety when inside the safe set, but also minimally modify the nominal control action to preserve safety when necessary. However, CBFs are often hand-designed, requiring an expert to tune the function for a given task specification. In addition, designing these CBFs requires tuning the exponential decay term, which governs the aggressiveness of the system's controller. Data-driven CBF methods \cite{liu2023safe, lavanakul2024safety, pmlr-v164-dawson22a} learn the control invariant set and associated control policy through supervised learning or reinforcement learning. Typically, these methods provide no formal guarantees on these learned approximations, but recent works use uncertainty quantification and other tools to provide formal or probabilistic guarantees \cite{kim2024learning, hu2024verificationneuralcontrolbarrier, 9303785, xiao2021barriernetsafetyguaranteedlayerneural}.
Moreover, as demonstrated in prior work \cite{tonkens2024patching, 9982058} and our findings, the learned CBF often fails to satisfy the necessary conditions to be a valid CBF, especially in the presence of control bounds.

\begin{wrapfigure}{r}{0.35\linewidth}
    \vspace{-6mm}
    \centering
    \includegraphics[width=\linewidth]{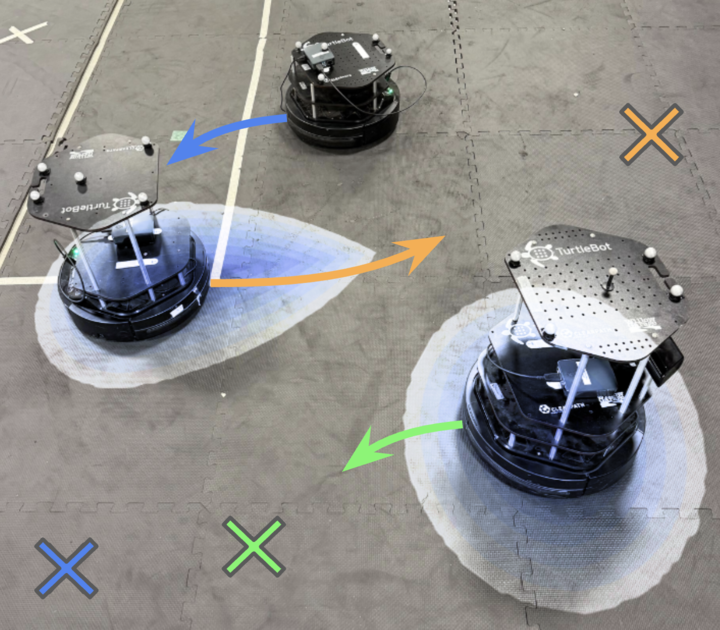}
    \caption{\textbf{Sample experiment setup.} Agents, endowed with a joint safety filter to avoid collisions, are independently attempting to reach their goals. The blue contours indicate the unsafe region for the blue agent at its heading.}
    \label{fig:frontpage}
    \vspace{-2mm}
\end{wrapfigure}

Hamilton-Jacobi reachability (HJR) analysis is a method that can be used for safety analysis that rigorously evaluates the set of states that may lead to failure under worst-case conditions by solving a partial differential equation (PDE) \cite{Bansal2017,borquez2024safety}. The result of the analysis is a value function, whose level sets and gradients inform the safe set and associated safe control policy. Recent work \cite{choi2021robust, tonkens2022refining} demonstrates that HJR can be adapted to directly compute CBF-like functions. This allows for the blending of the two approaches for applications in robotics and controls, merging the ease of use of CBFs with the rigorous guarantees and constructive nature of HJR. However, two issues remain: a) HJR relies on dynamic programming, which suffers from the curse of dimensionality, and b) the resulting CBF may not be differentiable everywhere. This differentiability issue is critical, as the gradient must exist to generate the set of safe actions at each state \cite{8796030}.

We propose a resolution to these two key issues by building upon DeepReach \cite{bansal2020deepreach}, a framework that merges HJR with physics-informed neural networks (PINNs). DeepReach computes a high-confidence solution to the HJR problem using the residual PDE error as the loss function. 
Interestingly, the computation requirements of DeepReach scale with the complexity of the underlying solution (rather than with the number of state variables), making it suitable for high-dimensional analysis. Moreover, recent works \cite{nakamura2023online, borquez2023parameter, jeong2025robotssuggestsafealternatives} have shown that various inputs to a Hamilton-Jacobi solution can be parameterized, including the control input, disturbance bound, as well as the cost function, allowing for online adaptation in uncertain and changing environments. A final benefit is that the learning error can be estimated via conformal prediction, which can subsequently be used to provide probabilistic safety assurances for the system \cite{lin2023generating}. 

In this paper, we adapt DeepReach to compute finite-time CBFs, yielding a unifying approach to the construction of CBFs for high-dimensional nonlinear systems with probabilistic safety guarantees. Specifically, we make the following contributions: 

     \textit{Smooth High-Dimensional CBFs with Bounded Control Inputs.} We introduce the Reachability Barrier Network (RBN), a learned CBF approximation using a modified version of DeepReach. Because DeepReach employs sinusoidal activation functions, we guarantee that the resulting CBF approximation is differentiable, making it readily suitable for synthesizing safe actions. Due to the underlying HJR-based backbone, the proposed approach allows us to synthesize CBF approximations for general nonlinear systems, while accounting for input constraints. 
     
     \textit{Adaptive Conservativeness Online.} CBFs are parameterized by an exponential decay rate $\gamma$ that tunes the aggressiveness of the controller. We propose to use a $\gamma$-parameterized RBN, through a linear time (i.e. $O(N)$) self-supervised learning process, allowing for post-training adaptation of the controller. 
     
     \textit{Probabilistic Safety Guarantees.} We provide probabilistic safety assurances to the RBN generated from conformal prediction tools.  Using these assurances, we augment the learned safe set to provide the desired probabilistic guarantees. We explore the relationship between the aggressiveness of the controller (via the decay rate $\gamma$) and the resulting probabilistic guarantees. 

We first compare our approach to the dynamic programming-based solution in a low-dimensional 3D Dubin's car setting. We then demonstrate the scalability of our approach in a higher-dimensional 9D multi-vehicle collision avoidance setting and compare it to a popular neural CBF architecture from \cite{pmlr-v164-dawson22a}. We validate our approach offline and online via conformal predictions and trajectory rollouts, respectively. Moreover, we analyze the robustness of our method compared to other baselines in a goal-seeking multi-vehicle collision avoidance hardware experiment.

\section{Background}\label{sec:background}
\subsection{Hamilton-Jacobi Reachability}

Consider a dynamical system with state $\state \in \mathbb{R}^{n_s}$, time horizon $\tint = [\tvar, \thor]$,  trajectory $\traj:\tint \to \mathbb{R}^{n_s}$, control input $u \in \cset \subset \mathbb{R}^{n_u}$,  control signal $\csig: \tint \to \cset$, and initial condition $\traj (t)=x$. 
Let the evolution of the system be defined by $\dot{\traj}  = \dyn ( \traj (\tvar), \csig (\tvar) )$, which we assume to be Lipschitz continuous with respect to $\traj$ and $\csig$, and continuous with respect to $t$. 

Consider a failure set $\failureset \subset \mathbb{R}^{n_\state}$, which the state may not enter, to be encoded by a Lipschitz continuous function $\safemargin:\mathbb{R}^{n_x} \to \mathbb{R}$ s.t. $\failureset := \{ \state \mid \safemargin(\state) < 0 \}$. We seek to solve $\csig$ that keeps $\traj$ out of $\failureset$ at all times in the horizon. 
Let the value then for any initial $(x,t)$ then be defined by \cite{Mitchell05},
\begin{equation}\label{eqn:cost}
    \valfunc(\state,\tvar) = \sup_{\csig \in \mathbb{U}} \min_{s\in [\tvar , \thor]} \safemargin(\traj(s)).
\end{equation}
Note, the zero level set of this value, $\mathcal{R}(t)\triangleq\{\state \mid V(\state,t) \le 0\}$, characterizes the set of initial $(\state,t)$ for which there exists a safe $\csig$ s.t. $\traj(\tau) \notin \failureset$, 
\begin{equation}\label{eqn:cost}
    \mathcal{R}(t) = \{ \state \mid \exists \csig \text{ s.t. } \traj(\tau) \notin \failureset, \tau \in \tint \},
\end{equation}
and is thus called the \textit{avoid tube}.
Notably, the value function is the solution to the HJ Variational Inequality \cite{Evans84, lygeros2004reachability},
\begin{equation}\label{eq:hjbvi}
    0 = \min \bigg\{ \safemargin(\state) - \valfunc(\state, \tvar),  
    \; \frac{\partial \valfunc}{ \partial t} + H(\state, \nabla_x V, \tvar) \bigg\}, \;\;
    \valfunc(\state, t_f) = \safemargin(\state)
\end{equation}
where $H(\state, p, \tvar) = \max_{\ctrl\in \cset} p \cdot \dyn(\state,\ctrl)$ is the Hamiltonian of the system. Moreover, the value function yields an optimal safe control policy for online control \cite{Evans84},
\begin{equation}
    \label{eq:safe_policy_case_1}
    u^*(x, t) =  \arg\max_{u\in U} \nabla_x V(x,t) \cdot \dyn(\state,\ctrl).
\end{equation}
This optimal control policy ensures that there is no change in safety over the time horizon, i.e. the value of the trajectory is constant over time. While safe, this may be overly conservative as it does not allow for \textit{any} decrease in safety, even when the system starts far from the failure set.  Alternatively, a least-restrictive hybrid controller can be applied \cite{hsu2023safety}: the optimal safe control policy activates only when the system is close to the safe boundary, and at all other times a separate performant controller is used. However, this switching-based control law often results in undesirable jerky behaviors and is vulnerable to numerical errors in the gradient of the value function -- or small approximation errors in the value itself \cite{borquez2024safety}.



\subsection{Control Barrier Functions}

Control barrier functions \cite{ames2019control} are a powerful tool for guaranteeing the safety of dynamic systems, ensuring that a system remains within a safe region indefinitely. 
A control barrier function by definition satisfies,
\begin{equation}\label{eq:CBF-constraint}
    \underset{u \in U}{\max} \; \nabla B(x) \cdot f(x, u) \geq -\alpha(B(x)), 
    \quad \forall x \in C, \;
    C = \{x \in \mathbb{R}^n : B(x) \geq 0 \},
\end{equation}
where $\alpha$ is a class $\kappa_{\infty}$ function, often chosen to be $\alpha(B(x))=\gamma B(x)$ (see \cite{hsu2023safety, ames2019control} for more details).
This condition ensures that any control input within the control bounds will cause a system to remain within the safe region.
Online, the resulting CBF may be used as a safety filter \cite{hsu2023safety},
\begin{equation} \label{eq:qp}
\begin{gathered}
u^*(x)=\arg \min _{u \in \mathcal{U}}\left\|u - u_{\text {nom }}(x)\right\|_2^2 \;
\text { s.t. \eqref{eq:CBF-constraint} holds.}
\end{gathered}
\end{equation}
Here $u_\text{nom}: \mathbb{R}^n \times \mathbb{R} \to \mathbb{R}^m$ represents a given nominal control law that might violate control limits and safety. 
For control-affine systems, \eqref{eq:qp} is a quadratic program, enabling real-time safety filtering. 

The main drawback of CBF-based approaches is the challenge of constructing a valid CBF. 
Often, CBFs are hand-designed for specific applications and systems. 
Some recent constructive methods have been proposed specifically for collision avoidance applications \cite{Squires2018ConstructiveCBF} and by using data-driven approaches \cite{Robey2020LearnCBFExpertDemonstrations,Wang2020LearningCBFHighRelDegree,Taylor2020LearningCBF,Srinivasan2020LearnCBFSupervisedML, Dawson2021CBFNN}. 
In many cases, these methods produce approximate CBFs that require significant parameter tuning \cite{nguyen2016optimal} and may not have formal guarantees, particularly when faced with bounded control inputs.


\subsection{Control Barrier Value Functions (CBVFs)}
In an effort to provide a constructive method for CBFs, \cite{choi2021robust} introduced the notion of a control barrier value function (CBVF),  a CBF-like function $\cbvf:\R^n \times \tint \rightarrow \R$, defined as
\label{ref:maximal-cbf}
\begin{equation}
\label{eq:mr-cbf-value}
    \cbvf(x, t):=  \sup_{\csig \in \mathbb{U}} \min_{s\in[t,\thor]} e^{\gamma(s-t)} \safemargin(\state(s)).
    \end{equation}

\noindent CBVFs can be computed via HJR by solving \eqref{eq:hjbvi} with the class $\kappa_{\infty}$ function, $\alpha(\cbvf(x, t)) = \gamma \cbvf(x, t)$,
\begin{equation}\label{eqn:cbvf_HJI_VI}
    0 = \min \bigg\{ \safemargin(\state) - \cbvf(\state, \tvar),
    \:\: 
    \frac{\partial \cbvf}{ \partial t} + H(x, \nabla_x \cbvf, t) + \gamma \cbvf(x, t) \bigg\}, \;\;
    \cbvf(\state, \thor) = \safemargin(\state).
\end{equation}


\noindent This allows for the construction of CBF-like functions over a finite time horizon with user-defined control bounds, disturbance bounds, and discount rate $\gamma$. This constructive approach has been used to refine CBF approximations from hand-tuned or data-driven approaches \cite{tonkens2022refining, tonkens2024patching}.  

However, there are two main challenges with the CBVF formulation.  First, the CBVF is not technically a CBF directly because the function may not be everywhere differentiable. This is problematic for the online CBF-QP controller ~\eqref{eq:qp}, which relies on the gradients of the function to maintain safety. The second challenge is that this technique for the construction of CBFs suffers from the aforementioned ``curse of dimensionality.''    
To be useful for many practical autonomous systems, the scalability of CBF construction must be pushed to handle higher-dimensional and more realistic models of systems and their operating environments.

\section{Learned Hamilton-Jacobi CBFs via \deepcbf s}
\subsection{Offline Learning Procedure}
\label{sec:learning}

We propose Reachability Barrier Networks (\deepcbf s) as a unifying method for the
construction of CBFs for high-dimensional nonlinear systems with probabilistic safety guarantees. Specifically, we propose modifying DeepReach \cite{bansal2020deepreach}, PINN used to solve HJ-PDEs, to compute a parameterized approximation of a CBF. 
The residual of \eqref{eqn:cbvf_HJI_VI} is employed as a loss $\mathcal{L}$ to learn the CBVF, 
\begin{align}\label{eq:loss_and_hamiltonian}
\mathcal{L}(\theta) &= \mathbb{E}_{x, t, \gamma} \left[
\bigg\Vert
\min\bigg\{
l(x) - V_\theta(x, t, \gamma),\;
\frac{\partial V_\theta}{ \partial t} + H(x, \nabla_x V_\theta, t) + \gamma V_\theta(x, t, \gamma)
\bigg\}
\bigg\Vert \right], \notag
\end{align}
where $\valfunc_\theta(x, t, \gamma) = l(x) + (t_f - t) \cdot \text{NN}_{\theta}(x, t, \gamma)$  and $\text{NN}_{\theta}$ is a neural network with sinusoidal activation functions and parameters $\theta$ \cite{sitzmann2020implicit}. The learned value $\valfunc_\theta$ is by definition infinitely differentiable and grid-free, overcoming the challenges of CBFs and CBVFs. Moreover, this approach is amenable to conformal corrections with probablistic assurances as we discuss in the following section.

To allow for online adaptation of the safety filter, we input the exponential decay rate $\gamma$ as a learned parameter to the model. Akin to a CBF, when $\gamma$ is high, the resulting RBN allows for more aggressive behavior, and as $\gamma$ approaches zero, the RBN becomes conservative, forcing safe behavior. 
Adding $\gamma$ to the state space increases the dimension by one, but the run-time cost is negligible in high-dimensions as neural nets do not suffer from the ``curse of dimensionality.''

\subsection{Probabilistic Guarantees via Conformal Prediction}
\label{sec:conformal}
After training, we use the conformal prediction-based method proposed in \cite{lin2024verification} to estimate the learning error and derive a probabilistic assurance on the learned CBF approximation. 
Namely, given a desired probability $\epsilon$ of safety, the method estimates an upper bound $\delta$ on the learned error such that the super-$\delta$ level of $V_\theta(x,t,\gamma)$ is safe at least with probability $1-\epsilon$,
\begin{equation} \label{eqn:probability_safety}
\mathbb{P}\left(V(x,t,\gamma) \leq 0 \mid V_\theta(x,t,\gamma) > \delta \right) < \epsilon 
\end{equation}

The learned value function $V_\theta$ is then shifted,
\begin{equation}
    V_\theta^\delta(x,t, \gamma) = V_\theta(x,t, \gamma) - \delta
\end{equation}
such that the super-$\delta$ level set becomes the new safe set with the associated $1-\epsilon$ probability of validity. Note, the gradients of the value function do not change and a robust learned value is recovered.

We note the $\gamma$ parameterization may also affect the probabilistic guarantee. As $\gamma$ varies, the gradients of $V_\gamma$ \cite{choi2021robust} are affected, which can both improve or degrade the learning performance and thus result in varying levels of $\delta$-correction. This is investigated in Fig.~\ref{fig:epsilon_volume_and_set_expansions} and Sec.~\ref{sec:results}.
In practice, one may tune $\gamma$ or $\delta$ to adapt the RBN in real time to meet the desired probabilistic constraints.  

\subsection{Online Control Synthesis}
Online, \deepcbf s can be used directly in the CBF-QP safety filter \eqref{eq:qp}, where the conservativeness of the filter can be varied via $\gamma$ in the \deepcbf s, $V_\theta(x,t,\gamma)$. 

\section{Simulation Results}\label{sec:results}
We first compare our approach to the dynamic programming-based CBVF from \cite{choi2021robust}. As this method is intractable in high dimensions, we compare the approaches in a 3D Dubin's car obstacle avoidance problem. We then scale our approach to a 9D multi-vehicle collision avoidance problem in both simulation and hardware, and compare with neural CBFs \cite{pmlr-v164-dawson22a}, a popular learning-based CBF, alongside a more standard pairwise collision avoidance approach. 

\subsection{Low Dimensional Model: 3D Dubin's Car}
The baseline model used to evaluate the performance of the \deepcbf s is a 3D Dubin's car model (note, the system becomes 4D with the parameterization of $\gamma$). The system is defined by $\dot x = v \cos(\theta)$, $\dot y = v \sin(\theta)$, $\dot \theta = u$, and $\dot{\gamma}=0$, where $x$ and $y$ are the euclidean coordinates, $\theta$ is heading, $v$ is a fixed velocity, our control is the angular velocity, $u \in [-1.1, 1.1]$. We train the sine-activated $3 \times 512$ PINN over $200K$ epochs with a learning rate of $2 \times 10^{-5}$ for approximately 4.5 hours on an A40 GPU. The low-dimensional ``ground truth'' CBVF is based on high-fidelity dynamic programming, using the HelperOC toolbox \cite{mitchell2007toolbox}. Additional parameters for our method and the HJR solution can be found in Appendix \ref{sec:appendix_training}.


\begin{figure}[h]
    \centering
    \begin{minipage}{0.67\columnwidth}
        \includegraphics[width=\linewidth]{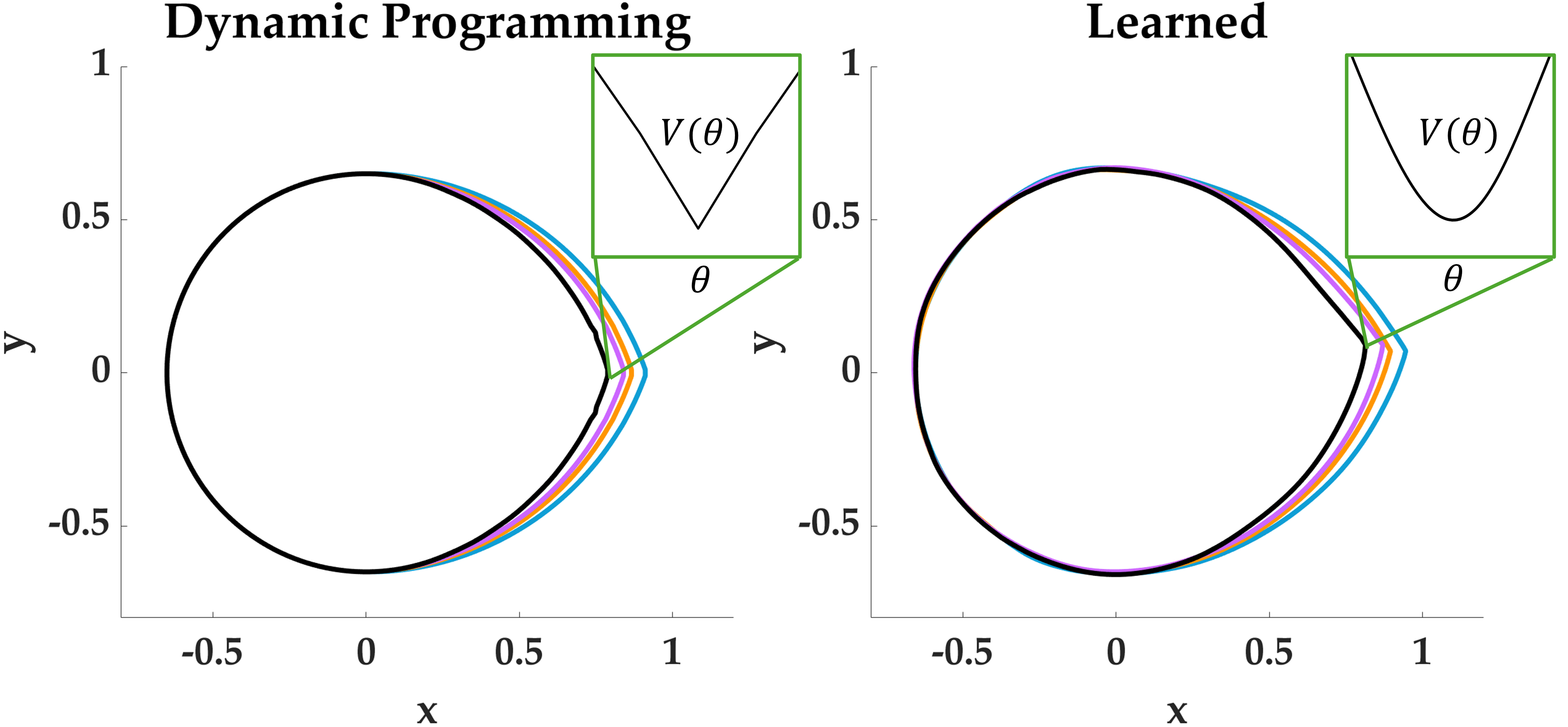}
    \end{minipage}%
    \hfill
    \begin{minipage}{0.3\columnwidth}
        \captionof{figure}{\textbf{Low-dimensional ground truth comparison.} The level set of the CBVF (left) and the RBN (right) are shown for $\theta=\pi$. Sets are plotted for $\gamma=\{0.0, 0.3, 0.5, 1.0\}$, shown in blue, orange, purple, and black, respectively. The insets (green) show the change in value over $\theta$, which exhibits the differentiability of the value function.}
        \label{fig:low_dim_set_comparison}
    \end{minipage}
    \vspace{-8mm}
\end{figure}


\begin{wraptable}{r}{0.4\linewidth}  
    \centering
    \caption{\textbf{Validation metrics of the learned CBF.} Sets are scored by discretizing the space and comparing against ground truth using IOU, FI, and FE.}
    \label{tab:3d_value_function_metrics}
    \begin{tabular}{ |c|c|c|c| }
        \hline
        $\gamma$ & IOU (\%) & FI (\%) & FE (\%) \\
        \hline
        0.0 & 96.85 & 2.51 & 0.64 \\
        0.3 & 97.57 & 1.90 & 0.53 \\
        0.5 & 97.77 & 1.81 & 0.42 \\
        1.0 & 97.87 & 1.89 & 0.24 \\
        \hline 
    \end{tabular}
    \vspace{-1em}
\end{wraptable}

\noindent \textbf{Comparison.} For values $\gamma \in \{0.0, \, 0.3, \, 0.5, \, 1.0 \}$, we compare the true and learned level sets and optimal policy trajectories at time $t=1$. Performance is evaluated using: a) the intersection-over-union (IOU) between the sets
; b) falsely included points (FI)
; and c) falsely excluded points (FE). 
We evaluate these metrics directly on the learned CBF approximation (without conformal expansion).


The results in Table~\ref{tab:3d_value_function_metrics} indicate that the learned value function closely approximates the dynamic programming-based solution. Fig.~\ref{fig:low_dim_set_comparison} shows slices of the \deepcbf and dynamic programming-based CBVF at different rates of $\gamma$ (note, we show the super-0.4 level sets, as the super-0 level sets are by definition identical across values of $\gamma$). The inset of the learned value function plotting $V(\theta)$ shows that our approach is everywhere differentiable (unlike CBVFs), allowing for more effective safety filtering via \eqref{eq:qp}.


\subsection{High Dimensional Model: 9D Multi-Vehicle Collision Avoidance}
The 9D multi-vehicle collision avoidance problem consists of three Dubin's cars seeking to avoid collision with one another. The system dynamics combine the individual models of the three Dubin's cars, with an additional parameterization of $\gamma$. Failure in simulation occurs when the distance between two vehicles falls below a collision radius of 0.25 meters. We train the sine-activated $3 \times 512$ PINN on 300K epochs with a learning rate of $2 \times 10^{-5}$. 

\noindent
\begin{minipage}{\linewidth}
    \vspace{1em}

    \centering\includegraphics[width=1\linewidth]{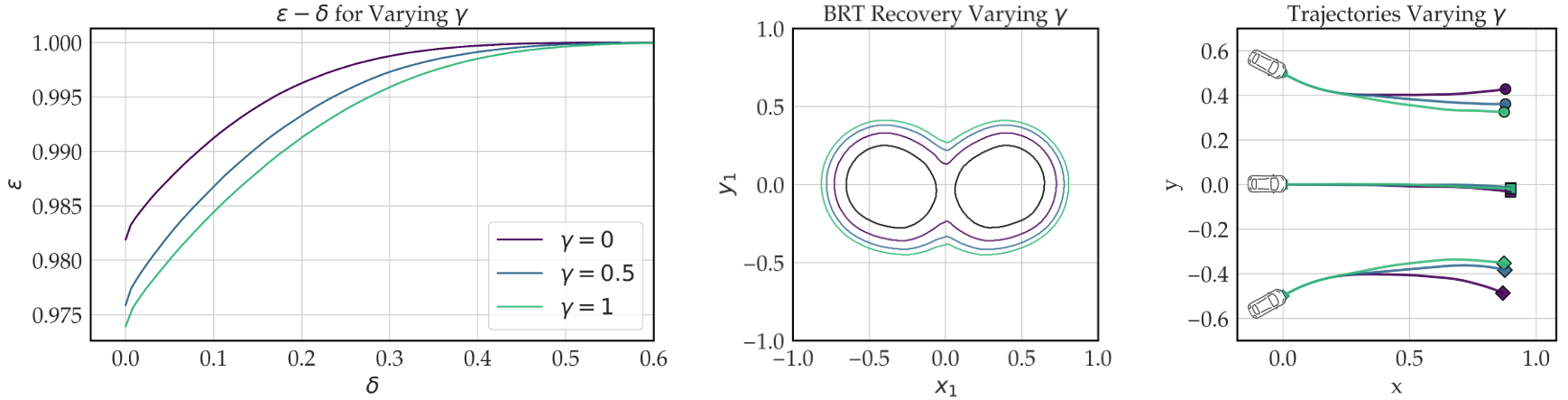}
    
    \captionof{figure}{\textbf{Conformal expansions of the 9D learned value functions.} (Left) Relationship between $\delta$ and probabilistic guarantee of safety ($\epsilon$) for varying $\gamma$. (Middle) Comparison of conformal expansions for $\gamma = {0, 0.5, 1}$. The uncalibrated zero-level set is included in black. (Right) Trajectory rollouts of the 9D multi-agent collision avoidance problem. Increasing $\gamma$ makes the control more aggressive and decreases the distance between agents. }
    \label{fig:epsilon_volume_and_set_expansions}
\end{minipage}

To derive probabilistic guarantees on the learned value function, we leverage conformal prediction \cite{lin2024verification} to expand the level sets of the value function. We sample $3M$ initial conditions and simulate trajectory rollouts using the learned policy. For each trajectory, we compare the final cost, which is the minimum distance between agents over the trajectory minus the collision radius, to the value assigned at the initial condition by the learned model. This yields a distribution of discrepancies between the learned and safety controller estimates. With this distribution, a confidence $\epsilon$ with a corresponding violation $\delta$ is used to determine that all learned trajectories will have the correct safety representation (i.e. value sign) with confidence $\epsilon$. Fig. \ref{fig:epsilon_volume_and_set_expansions} (left, middle) shows that an increase in the aggressiveness of our controller (i.e. an increase in gamma) requires a larger expansion of the safe set for the same probabilistic assurance. This follows intuitively, since Fig.~\ref{fig:epsilon_volume_and_set_expansions} (right) shows a reduction in minimum distance as $\gamma$ increases, equivalent to more aggressive controls. We choose $\epsilon \in [0.985, 0.995]$ to prevent outliers from influencing the recovered set to unnecessarily large proportions while also rigorously maintaining safety. 

\begin{wrapfigure}{r}{0.65\linewidth}
    \centering
    \includegraphics[width=\linewidth]{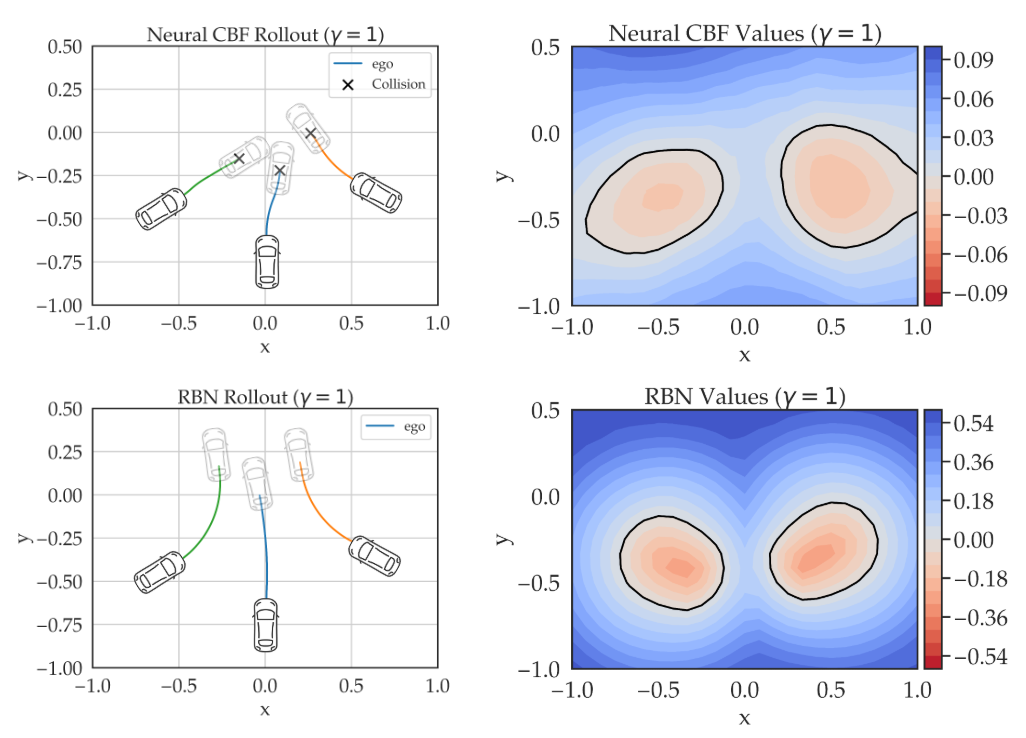}
    \caption{\textbf{Rollouts and contours.} (Left) Sample rollouts and their corresponding initial condition level sets (Right) for neural CBFs (Top) and our method (Bottom).}
    \label{fig:9d_rollout_comparisons}
    \vspace{-14mm}
\end{wrapfigure}

\textbf{Comparison.} We compare our \deepcbf~method to the neural CBF approach proposed by \cite{pmlr-v164-dawson22a}. For a single training run, both methods require approximately 7 hours of training on A40 GPUs. However, unlike \deepcbf, the neural CBF approach requires retraining the network for each desired gamma value. We use the false positive rate (FPR), and the false negative rate (FNR) to determine our model's performance. A false positive occurs when the initial condition is in the learned safe set (i.e. $V_\theta(x,t,\gamma) \ge 0$) but the rollout results in collision. The negation is true for false negatives. 
We also include the correctly characterized (CC) percentage, given by $1 - \text{FPR} - \text{FNR}$. Table \ref{tab:combined_rollout_metric_comparison} shows that our method best preserves safety for the multi-vehicle collision avoidance problem with or without a QP. While our expanded method yields the highest FNR, this is to be expected as the value function becomes more conservative and does not affect the true success of the rollouts. Fig. \ref{fig:9d_rollout_comparisons} shows a sample state where the neural CBF is unable to maintain safety for a region our method is verifiably safe. 

\vspace{-6mm}

\begin{table*}[t]
    \centering
    \scriptsize 
    \setlength{\tabcolsep}{4pt} 
    \renewcommand{\arraystretch}{1.3} 
    \begin{tabular}{|c|c|c|c|c|c|c|c|c|c|c|}
        \hline
        \multirow{2}{*}{Setting} & \multirow{2}{*}{Metric} & \multicolumn{3}{c|}{Neural CBF \cite{pmlr-v164-dawson22a}} & \multicolumn{3}{c|}{\deepcbf~(Ours)} & \multicolumn{3}{c|}{Expanded \deepcbf~(Ours)} \\ \cline{3-11}
        & & $\gamma=0.0$ & $\gamma=0.5$ & $\gamma=1.0$ & $\gamma=0.0$ & $\gamma=0.5$ & $\gamma=1.0$ & $\gamma=0.0$ & $\gamma=0.5$ & $\gamma=1.0$ \\ \hline

        \multirow{3}{*}{QP Rollout} 
        & FPR (\%) & 28.4 & 28.2 & 30.8 & \textbf{0.8} & \textbf{1.4} & \textbf{2.2} & \textbf{0.0} & \textbf{0.0} & \textbf{0.0} \\
        & FNR (\%) & 1.8 & 1.6 & 1.6 & 0.0 & 0.0 & 0.0 & 6.8 & 12.0 & 16.0 \\
        & \textbf{CC (\%)} & 69.8 & 70.2 & 67.6 & 99.2 & 98.6 & 97.8 & 93.2 & 88.0 & 84.0 \\ \hline

        \multirow{3}{*}{\shortstack{Learned Policy\\Rollout}} 
        & FPR (\%) & 27.0 & 28.2 & 30.0 & \textbf{0.0} & \textbf{0.0} & \textbf{0.0} & \textbf{0.0} & \textbf{0.0} & \textbf{0.0} \\
        & FNR (\%) & 2.2 & 2.6 & 1.4 & 0.4 & 0.4 & 0.4 & 7.4 & 10.0 & 11.2 \\
        & \textbf{CC (\%)} & 70.8 & 69.2 & 68.6 & 99.6 & 99.6 & 99.6 & 93.6 & 90.0 & 88.8 \\ \hline
    \end{tabular}
    \caption{\textbf{Simulated rollout metrics} across $\gamma$ values for Neural CBF, \deepcbf, and Expanded \deepcbf in both QP Rollouts and Rollouts using only the Learned Safe Policy (i.e. no nominal control).}
    \label{tab:combined_rollout_metric_comparison}
    \vspace{-2.5em} 
\end{table*}


 


\noindent

\section{Hardware Results}

The hardware setup is designed to represent an autonomous multi-vehicle context, and mimics the simulations. The Dubin's car model is used to drive TurtleBots, operating at a constant linear velocity of $v=0.4$ m/s, with control inputs $u \in [-1.1, 1.1]$ rad/s. We use a motion capture system to assume perfect state estimation; a detailed explanation can be found in Appendix \ref{sec:appendix_hardware}. During each 60-second trial, we solve the CBF-QP (\ref{eq:qp}) to generate safe control inputs that let each agent pursue its assigned goal independently while jointly avoiding collisions—defined as any time the distance between two agents falls below 0.4 m. The goals are defined by targets with a radius of 0.3 m. When an agent reaches its goal, a new goal is assigned, independent of the other agents. If a collision occurs, all agents are instructed to stop and turn away from each other until they are safe to avoid chained collisions. A simple proportional controller nominally controls the angular velocity, directing each agent towards its goal region. However, in some cases, the safety constraints produce deadlock scenarios in which agents are unable to make progress and ultimately collide. We address this issue using the resolution strategy described in Appendix \ref{sec:appendix_hardware}.

As solving the full 9D multi-vehicle collision avoidance game via dynamic programming is infeasible, we first compare our method to a two-player dynamic-programming solution. Online, the two-player solution is used in a decentralized, pairwise manner: each agent computes its own CBF relative to its nearest neighbor, where the ego agent assumes the other agent acts cooperatively. While this two-player baseline cannot fully capture higher-order multi-agent interactions, it serves as the strongest available solution that can be computed exactly. This comparison thus highlights both the practical limitations imposed by the curse of dimensionality and the benefits of scalable, learning-based approaches such as RBNs. 
Our principal comparison is to neural CBFs \cite{pmlr-v164-dawson22a}, as both methods are trained to address the same 9D multi-vehicle collision avoidance problem. All training hyperparameters can be found in Appendix \ref{sec:appendix_training}. 

\begin{table*}[t]
    \centering
    \scriptsize
    \setlength{\tabcolsep}{6pt}
    \renewcommand{\arraystretch}{1.3}
    \begin{tabular}{|c|ccc|ccc|ccc|}
        \hline
        \multirow{2}{*}{\textbf{Controller}} 
        & \multicolumn{3}{c|}{$\gamma = 0.0$} 
        & \multicolumn{3}{c|}{$\gamma = 0.5$} 
        & \multicolumn{3}{c|}{$\gamma = 1.0$} 
        \\ \cline{2-10}
        & Collisions & In-Dist & Goals & Collisions & In-Dist & Goals & Collisions & In-Dist & Goals \\
        \hline
        \deepcbf (ours)   & 0.2  & \textbf{0.0}  & 12.6 & 0.4  & \textbf{0.2}  & \textbf{29.2}  & 1.8  & 0.8 & \textbf{30.6}   \\

        Two-player (coop) & \textbf{0.0}  &  \textbf{0.0}  & 5.4  & \textbf{0.2}  &  \textbf{0.2}  & 20.6 & 2.2  & 2.2 & 24.4   \\
        
        
        Neural CBF    & 2.6  & 1.4  & \textbf{14.0} & 4.2  & 3.0  & 19.6  & 6.0  & 4.4 & 16.0   \\
        Nominal & 7.4  & 7.4   & 20.8 & 7.4  & 7.4   & 20.8  & 7.4  & 7.4  & 20.8   \\
        
        \hline
    \end{tabular}
    \caption{\textbf{Hardware comparison} of total collisions, in-distribution collisions, and goals reached for \deepcbf~(ours), neural CBF, and HJR-based controllers across different values of $\gamma$. All metrics are reported per minute and are averaged over 5 minutes of hardware simulation. }
    \label{tab:gamma_sweep_results}
    \vspace{-5mm}
\end{table*} 

To evaluate the safety and efficiency of each approach, we compare the success of reaching goals and the number of collisions incurred as the primary metrics. We focus our evaluation on in-distribution collisions, as the learned methods are only trained within a prescribed domain: specifically, $x \in [-1, 1]$, $y \in [-1, 1]$, and $\theta \in [-\pi, \pi]$. However, we also report the total number of collisions to assess how well learning-based approaches generalize beyond their training distribution.

\noindent \textbf{Comparison to pairwise collision avoidance.} Our in-distribution collision rate, as shown in Table \ref{tab:gamma_sweep_results}, is comparable to that of the cooperative two-player solution when $\gamma=0.0$ or $0.5$. However, since increasing $\gamma$ reduces conservativeness, more three-agent interactions occur when $\gamma=1.0$, resulting in 2.8x the in-distribution collision rate for the two-player cooperative solution compared to our method. Furthermore, our method achieves 1.4x the goal completion rate of the two-player cooperative solution at $\gamma=0.5$, and 1.3x at $\gamma=1.0$.
We hypothesize that increasing the number of agents will correlate with a higher relative score for our method compared to the HJR solution since the pairwise solution are unable to safely handle multi-agent interactions. 

\noindent \textbf{Comparison to neural CBFs.} We attribute the subpar performance of neural CBFs in part to difficulties in constructing an effective boundary region between safe and unsafe sets. The neural CBFs require predefining safe and unsafe regions, whereas our method only requires instantiating the boundary function. This allows our method to be less complex to implement and train. Ultimately, the RBNs on average incur 5.5x fewer collisions and reach 1.9x as many goals as the neural CBFs.

Overall, our method performs safely even when using highly aggressive controls at $\gamma=1.0$, whereas increasing the aggressiveness for other methods compromises their safety. Our controller also exhibits the lowest deviation from the nominal objective, consistently achieving the highest goal completion rates. 
This trend is also illustrated in Fig.~\ref{fig:all_methods_hw}, where, under induced three-agent collisions, our controller most effectively avoids collisions. 

\noindent
\begin{minipage}{1.0\linewidth}
    \centering
    \includegraphics[width=\linewidth]{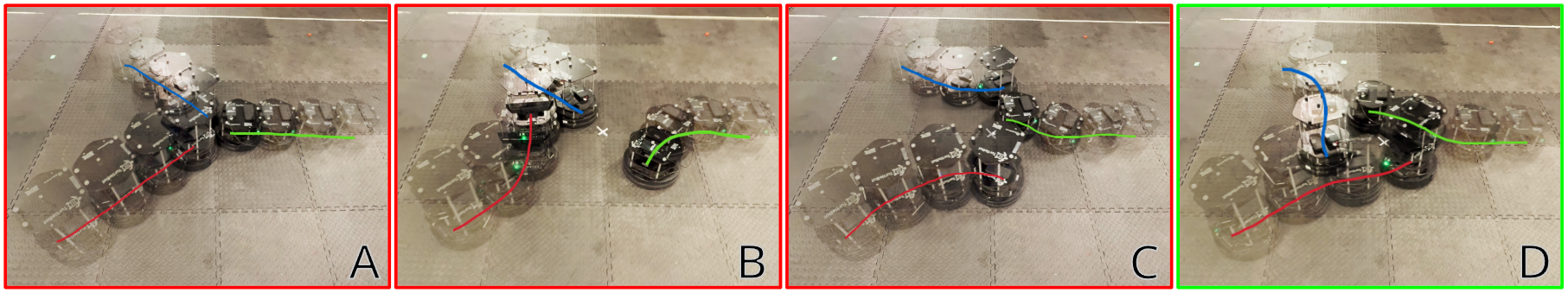}
    \captionof{figure}{\textbf{Comparison of the controllers in hardware} (the robots become less transparent as time progresses). A) Nominal controller.  B) neural CBF controller. C) HJ cooperative pairwise controller. D) RBNs (our method). Our method is the only one capable of safely navigating this scenario. }
    \label{fig:all_methods_hw}
\end{minipage}


\section{Conclusion}\label{sec:conclusions}
In this paper, we introduce \deepcbf s, a method for computing a $\gamma$-parameterized control barrier function, and discuss the insights of the probabilistic guarantees for varying $\gamma$. We build intuition through a low-dimensional navigation example and demonstrate the scalability of our work through a high-dimensional multi-vehicle collision avoidance example. We demonstrate the efficacy of our method through hardware experiments, where we achieve high success rates relative to other methods. 

\newpage
\section{Limitations}
\textbf{\textit{Computational scalability}}. While RBNs scale better than dynamic programming, training still requires significant compute (e.g., 7 hours on an A40 GPU for the 9D system). For very high-dimensional systems, training cost could become significant. We aim to explore sampling-efficient or warmstarting approaches \cite{sharpless2025linearsupervisionnonlinearhighdimensional} to address this problem. Note, the neural CBF method also suffers from the same compute requirements. 

\textbf{\textit{In-distribution safety}}. The probabilistic guarantee generated by conformal prediction weakens as $\gamma$ increases; we aim to address this limitation through online adaptation of \deepcbf~via a tradeoff between aggressive behavior arising from $\gamma$ variance and safety assurances (i.e. corresponding probabilistic guarantees).

\textbf{\textit{Probabilistic assurances rely on the exchangeability assumption}}. If the system encounters scenarios not well covered by the training or calibration data, safety may not be assured. Future works will include leveraging uncertainty/risk quantification methods \cite{10.1007/978-3-030-28619-4_10} to handle out-of-distribution scenarios.

\textbf{\textit{Modeling Assumptions}}. The approach assumes the system model is known (e.g., Dubin's car) and the constraint set is also known a priori. We are excited to address this in future work via reducing order modeling \cite{10155871}, where disturbances in the system dynamics act as the model mismatch error. 

\textbf{\textit{State Estimation Assumption}}. Our current method assumes perfect state estimation through a motion capture arena. On the contrary, robots in the real world rely on imperfect state estimation. In future works, we will explore various sensors (e.g. LiDAR, RGB Camera), as well as accounting for state estimation errors via disturbance modeling \cite{lin2024one}.

\clearpage

\bibliography{corl_2025/references}

\newpage
\section{Appendix}\label{sec:appendix}
\begin{figure}[h]
    \centering
    \includegraphics[width=1.0\linewidth]{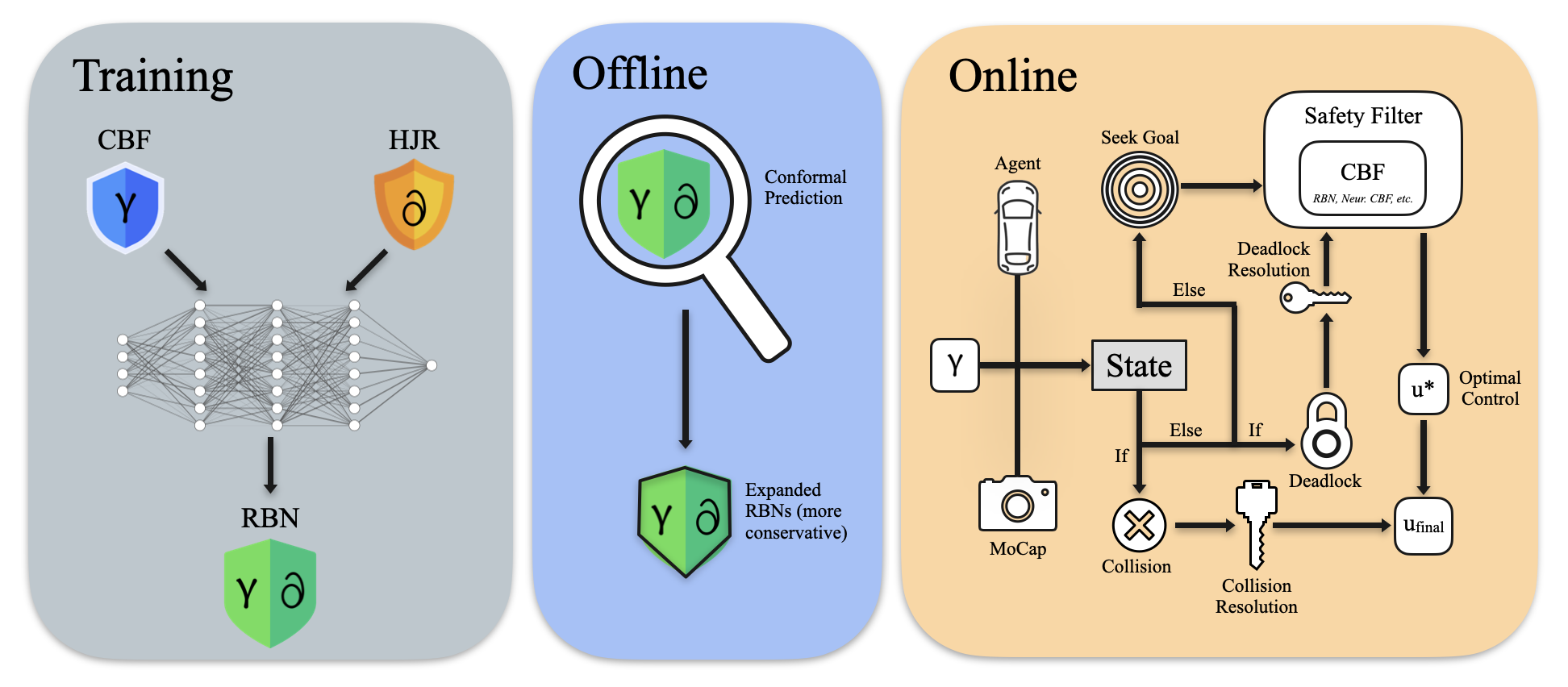}
    \caption{\textbf{Method overview.} During training, we integrate CBFs with HJR to obtain a $\gamma$-parameterized RBN. We subsequentally obtain probablistic guarantees post-training. Online, we use RBN as a safety filter, where we account for both collision resolution and deadlock detection.}
    \label{fig:method}
\end{figure}

Detailed explanations of the training process and conformal prediction are provided in Sections~\ref{sec:learning} and~\ref{sec:conformal}, respectively. For the full online control pipeline illustrated in Figure~\ref{fig:method}, we begin by obtaining the agent states from the motion capture system. The parameter $\gamma$ is appended to the state; this can be done at any point prior to passing the values to the safety filter.

We first check for collisions. If any are detected, we zero the agents' velocities and apply turning controls to separate them. Once agents are no longer in collision (i.e., at least $0.4 \text{m}$ apart), we restore their velocities (e.g., $0.4 \text{m/s}$) and resume the nominal policy.

Following collision resolution, we apply either the nominal goal-seeking policy or deadlock-resolution control, depending on whether a deadlock is detected. These controls are then passed through the RBN-based safety filter, which enforces safety via a CBF-QP. Both the nominal and deadlock policies are described in Appendix~\ref{sec:appendix_hardware}.

\subsection{Training Details}
\label{sec:appendix_training}

We list the parameters used to train the HJ pairwise solution, the neural CBF solutions, and our RBN solution, along with the domain parameters for the Dubin's systems. We keep parameters consistent across methods to ensure fairness during evaluation.  

\begin{table}[h!]
\centering
\scriptsize 
\begin{minipage}[t]{0.33\textwidth}
    \centering
    \begin{tabular}{l r}
    \toprule
    \textbf{Parameter} & \textbf{Value} \\
    \midrule
    X & $[-1, 1]$ m \\
    Y & $[-1, 1]$ m \\ 
    $\Theta$ & $[-\pi, \pi]$ rad \\ 
    $v$ & 0.6 m/s \\
    $\dot{\theta} \; (u)$ & $[-1.1, 1.1]$ rad/s \\
    Collision Radius & 0.4 m \\

    \bottomrule
    \end{tabular}
    \vspace{0.5em}
    \caption{System Parameters}
\end{minipage}
\end{table}

We restrict the state domains so that the learned methods can be trained with high fidelity in a reasonable amount of time. We also limit our angular and linear velocities so that they do not exceed the maximum physical capabilities of the TurtleBots. 

\begin{table}[h!]
\centering
\scriptsize 

\begin{minipage}[t]{0.33\textwidth}
    \centering
    \vspace{0pt}
    \begin{tabular}{l r}
    \toprule
    \textbf{Hyperparameter} & \textbf{Value} \\
    \midrule
    Grid Size & $80 \times 80 \times 60$ \\
    Solver Accuracy & High \\
    \bottomrule
    \end{tabular}
    \vspace{0.5em}
    \caption*{(a) Two-Player HJR: Grid}

\end{minipage}
\begin{minipage}[t]{0.33\textwidth}
\centering
\vspace{0pt}
\begin{tabular}{l r}
\toprule
\textbf{Hyperparameter} & \textbf{Value} \\
\midrule
Hidden Layers & 3 \\
Hidden Layer Size & 512 \\
Epochs & 51 \\
Learning Rate & $5 \times 10^{-4}$ \\
CBF $\lambda \; (\gamma) $ & $[0.0, 0.5, 1.0]$ \\
Relaxation Penalty & 200 \\
Controller Period & 0.01 s \\
Learn Shape Epochs & 21 \\
Scale Parameter & 10.0 \\
Use ReLU & True \\
Trajectories / Episode & 50 \\
Trajectory Length & 300 \\
Fixed Samples & 40,000 \\
Max Points & 50,000 \\
Validation Split & 0.1 \\
Batch Size & 64 \\
Boundary Quota & 0.5 \\
Unsafe Quota & 0.4 \\
Safe Quota & 0.0 \\
\bottomrule
\end{tabular}
\vspace{0.5em}
\caption*{(b) Neural CBFs}
\end{minipage}
\begin{minipage}[t]{0.33\textwidth}
\centering
\vspace{0pt}
\begin{tabular}{l r}
\toprule
\textbf{Hyperparameter} & \textbf{Value} \\
\midrule
$\gamma$ & $[0, 1]$ \\
Min With & Target \\
Sample Count & 65,000 \\
Pretrain & True \\
Pretrain Epochs & 20,000 \\
Time Range & $[0.0, 1.0]$ \\
Counter Start & 0 \\
Counter End & -1 \\
Source Samples & 1,000 \\
Target Samples & 0 \\
Activation & Sine \\
Model Mode & MLP \\
Hidden Layers & 3 \\
Hidden Layer Size & 512 \\
DeepReach Model & Exact \\
Batch Size & 1 \\
Learning Rate & $2 \times 10^{-5}$ \\
Epochs & 300,000 \\
Gradient Clipping & 0.0 \\
Adjust Relative Gradients & True \\
Dirichlet Loss Divisor & 1.0 \\
\bottomrule
\end{tabular}
\vspace{0.5em}
\caption*{(c) RBNs (DeepReach)}
\end{minipage}

\caption{Training and implementation parameters for each method.}
\label{tab:method_hyperparams}
\vspace{-6mm}
\end{table}

\textit{HJ Pairwise}.
We compute backward reachable tubes using dynamic programming over a discretized state space.
For the two-player setting, we use the \texttt{Air3D} dynamics model, which captures the relative state between two evaders in polar coordinates relative to the ego agent.
The system is described by the state $z = (x, y, \psi)$, where $(x, y)$ is the relative position between two evaders, and $\psi$ is the relative heading.
The avoid set, which represents the non-ego evader, is a circular region of radius 0.4m centered at the origin.
We solve the HJ PDE on a 3D grid of size $80 \times 80 \times 60$ with periodic boundary conditions in the angular dimension.
We use high-accuracy solver settings and compute the reachable tube over a time horizon of 1.0 seconds.

\textit{Neural CBFs.} All Neural CBF hyperparameters were selected empirically after evaluation on validation rollouts. We intentionally keep the size of the network and the sampling points to be the same as our RBN method. Another key design choice was how we defined safe and unsafe regions in the state space, which implicitly defines a boundary region where transitions occur. 

As mentioned in \cite{pmlr-v164-dawson22a}, a non-zero boundary is required to enable smooth transitions between the safe and unsafe sets, which helps avoid gradient discontinuities and training instability. we experimented with various settings. We chose to keep the boundary function consistent with what we use in our RBN method (i.e. min distance function). Specifically, we define the \textit{unsafe region} as the set of states where the minimum distance between agents is less than the collision radius (0.4 m), and the \textit{safe region} as the set where the minimum inter-agent distance exceeds the collision radius by at least 0.2 meters. The boundary region is thus the band between these two thresholds.

\textit{RBNs}. We adapt the hyperparameters from \cite{bansal2020deepreach}. We first pretrain the boundary condition for 20k epochs, ensuring that the terminal condition is properly learned. Then, we uniformly sample 65,000 points across the state space per epoch. We sequentially increase the time-horizon samples uniformly across epochs. To keep the PDE and boundary loss consistent, we adjust relative gradients between the two losses.

\subsection{Hardware Implementation Details} \label{sec:appendix_hardware}
\begin{wrapfigure}{r}{0.35\linewidth}
    \centering
    \includegraphics[width=\linewidth]{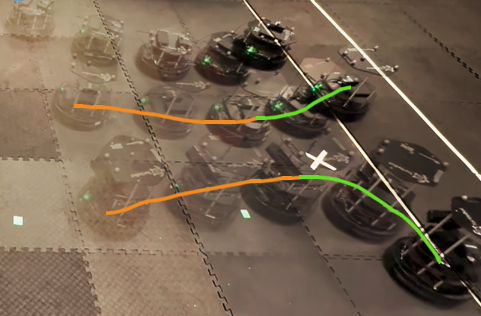}
    \caption{\textbf{Deadlock} occurrence (orange) and resolution (green).}
    \label{fig:deadlock}
\end{wrapfigure}

\textit{Deadlock}. We find that deadlock occurs in our long time-horizon multi-vehicle collision avoidance experiment for all of the CBF methods we test. The nominal control of deadlocked agents drive them towards unsafe regions, preventing them from achieving their goals. More importantly, deadlocked agents eventually exit the state distribution on which they were evaluated, leading to collisions. To accommodate, we replace the original goal-seeking nominal control of deadlocked agents with instructions to turn away from each other. We classify deadlock when two agents maintain similar headings:
\begin{equation}
    \mathcal{D}_{N,M} = |\theta_N - \theta_M| < \delta
\end{equation}
where $\mathcal{D}$ is the deadlock check, $N$ and $M$ is a unique pair of agents, and $\delta$ is the deadlock threshold. If $\mathcal{D}$ is true for $T$ consecutive timesteps, we consider the system to be in a deadlock. In practice, we empirically determined $T = 100$ (1 second) to consistently detect deadlock while minimally interfering with the nominal and safe controllers. As seen in Fig. \ref{fig:deadlock}, once deadlock is detected, we apply a corresponding deadlock resolution controller, defined as:
\begin{align}
u_{\mathcal{D},N} &= -k \cdot \text{sgn}\left( \cos(\theta_N) (y_M - y_N) - \sin(\theta_N) (x_M - x_N) \right) \\
u_{\mathcal{D},M} &= -k \cdot \text{sgn}\left( \cos(\theta_M) (y_N - y_M) - \sin(\theta_M) (x_N - x_M) \right)
\end{align}

Here, $u_{\mathcal{D}}$ is shown for each agent, sgn represents a sign function, and $k$ is the magnitude of the controller. To determine the direction each agent should turn towards to resolve deadlock, we take the sign of the cross product between the heading vector and the vector between the agents. We set $k=3$ during all experiments. We apply $u_{\mathcal{D}}$ repeatedly for $100$ timesteps to ensure that the robots escape deadlock and do not fall back into deadlock again. Note that $u_{\mathcal{D}}$ is passed through the safety filter to also ensure safety during deadlock resolution. A visualization of deadlock resolution is shown in \ref{fig:deadlock}.

\textit{Nominal policy}. For our nominal policy, we correct the angle error between each agent and their respective goals. This is denoted as:
\begin{align}
\theta^*_i &= \arctan\left( y^*_i - y_i,\ x^*_i - x_i \right) \\
\omega_i &= \arctan\left( \sin(\theta^*_i - \theta_i),\ \cos(\theta^*_i - \theta_i) \right)
\end{align}

where $i$ denotes each agent, $\theta^*_i$ denotes the target angle for each agent, and $\omega_i$ represents the nominal control (e.g. correction term).

\textit{Sim2Real}. Although our simulations are based on the Dubin’s car model, the physical implementation on TurtleBots is non-trivial because the Dubin’s car assumes instantaneous control input actuation, which is not the case when it comes to TurtleBots. To account for delays in control actuation, we implement a closed-loop PID controller that aligns each TurtleBot's physical motion with a simulated reference trajectory generated by our policy. Specifically, the PID controller adjusts linear and angular velocities such that the robot closely follows the true Dubin’s trajectory computed at each timestep.

\end{document}